\documentclass[twoside,11pt]{article}

\usepackage{hyperref}
\hypersetup{
    colorlinks=true,
    citecolor=cyan,
    linkcolor=cyan,
    urlcolor=cyan
}

\usepackage{refstyle}
\usepackage{jair, theapa, rawfonts}

\usepackage{xcolor}
\usepackage{times}
\usepackage{soul}
\usepackage{url}
\usepackage[utf8]{inputenc}
\usepackage{graphicx}
\usepackage{amsmath}
\usepackage{amsthm}
\usepackage{booktabs}
\usepackage[ruled]{algorithm2e}
\usepackage{color}
\usepackage{amsfonts}
\usepackage{cancel}
\usepackage{pifont}
\usepackage{amsmath}
\usepackage{graphicx}
\usepackage{multirow}
\usepackage{amssymb}

\theoremstyle{definition}

\firstpageno{1}

\begin{document}

\title{Pretraining in Deep Reinforcement Learning: A Survey}

%\author{}
\author{\name Zhihui Xie  \email fffffarmer@sjtu.edu.cn \\
      \addr Shanghai Jiao Tong University
      \AND
      \name Zichuan Lin \email zichuanlin@tencent.com \\
      \addr Tencent
      \AND
      \name Junyou Li \email junyouli@tencent.com \\
      \addr Tencent
      \AND
      \name Shuai Li \email shuaili8@sjtu.edu.cn \\
      \addr Shanghai Jiao Tong University
      \AND
      \name Deheng Ye \email dericye@tencent.com \\
      \addr Tencent }
      %\AND
      %\name Tengyu Ma \email tengyuma@stanford.edu \\
      %\addr Stanford University}

% For research notes, remove the comment character in the line below.
% \researchnote

\maketitle

% For thicker \hline
\makeatletter
\def\thickhline{%
  \noalign{\ifnum0=`}\fi\hrule \@height \thickarrayrulewidth \futurelet
   \reserved@a\@xthickhline}
\def\@xthickhline{\ifx\reserved@a\thickhline
               \vskip\doublerulesep
               \vskip-\thickarrayrulewidth
             \fi
      \ifnum0=`{\fi}}
\makeatother

\newlength{\thickarrayrulewidth}
\setlength{\thickarrayrulewidth}{2\arrayrulewidth}

% Set cross marks
\newcommand{\cmark}{\ding{51}}%
\newcommand{\xmark}{\ding{55}}%
% Comments
\newcommand{\zhihui}[1]{{\color{green}[#1]}}

% Components
\newcommand{\encoder}{{\phi\left(o\right)}}
\newcommand{\policy}{{\pi_\theta\left(a | s\right)}}
\newcommand{\cpolicy}{{\pi\left(a | s, z\right)}}
\newcommand{\controller}{{\pi(z \mid s)}}
\newcommand{\offdata}{{\mathcal{D}}}

\begin{abstract}
    The past few years have seen rapid progress in combining reinforcement learning (RL) with deep learning.
    Various breakthroughs ranging from games to robotics have spurred the interest in designing sophisticated RL algorithms and systems.
    However, the prevailing workflow in RL is to learn \textit{tabula rasa}, which may incur computational inefficiency.
    This precludes continuous deployment of RL algorithms and potentially excludes researchers without large-scale computing resources.
    In many other areas of machine learning, the pretraining paradigm has shown to be effective in acquiring transferable knowledge, which can be utilized for a variety of downstream tasks.
    Recently, we saw a surge of interest in \textit{Pretraining for Deep RL} with promising results.
    However, much of the research has been based on different experimental settings.
    Due to the nature of RL, pretraining in this field is faced with unique challenges and hence requires new design principles.
    In this survey, we seek to systematically review existing works in pretraining for deep reinforcement learning, provide a taxonomy of these methods, discuss each sub-field, and bring attention to open problems and future directions.
    
    % \zhihui{Refinement ends here.}
    % Pretraining has become a \textit{de facto} paradigm for natural language processing and computer vision.
    % Recently, the field of deep reinforcement learning (DRL) saw a similar surge of interest in leveraging abundant source data for pretraining, which provides promising solutions to build generalist agents.
    % These methods aim at learning general prior knowledge from either online reward-free environments or offline data, with the hope to facilitate faster adaptation or better generalization.
    % In this survey, we take a review of pretraining in DRL.
    % We advocate for special cares on the nature of RL (objectives, architectures, and benchmarks).
\end{abstract}

\section{Introduction}

Reinforcement learning (RL) provides a general-purpose mathematical formalism for sequential decision-making \shortcite{sutton2018reinforcement}.
By utilizing RL algorithms together with deep neural networks, various milestones in different domains have achieved superhuman performances via optimizing user-specified reward functions in a data-driven manner~\shortcite{silver2016mastering,akkaya2019solving,vinyals2019grandmaster,ye2020towards,ye2020mastering,ye2020supervised,chen2021heroes}.
As such, we have seen a growing interest recently in this research direction.

However, while RL has been proven effective at solving well-specified tasks, the issue of \textit{sample efficiency}~\shortcite{jin2021bellman} and \textit{generalization}~\shortcite{kirk2021survey} still hinder its application to real-world problems.
In RL research, a standard paradigm is to let the agent learn from its own or others’ collected experience, usually on a single task, and to optimize neural networks \textit{tabula rasa} with random initializations.
For humans, in contrast, prior knowledge about the world contributes greatly to the decision-making process.
If the task is related to previously seen tasks, humans tend to reuse what has been learned to quickly adapt to a new task, without learning from exhaustive interactions from scratch. 
Therefore, as compared to humans, RL agents usually suffer from great data inefficiency~\shortcite{kapturowski2022human} and are prone to overfitting~\shortcite{zhang2018dissection}.

Recent advances in other areas of machine learning, however, actively advocate for leveraging prior knowledge built from large-scale pretraining.
By training on broad data at scale, large generic models, also known as \textit{foundation models}~\shortcite{bommasani2021opportunities}, can quickly adapt to various downstream tasks.
This pretrain-finetune paradigm has been proven effective in areas like computer vision~\shortcite{chen2020simple,he2020momentum,grill2020bootstrap} and natural language processing~\shortcite{devlin2018bert,brown2020language}.
However, pretraining has not yet had a significant impact on the field of RL.
Despite its promise, designing principles for large-scale RL pretraining faces challenges from many sources: 1) the diversity of domains and tasks; 2)
the limited data sources; 3) the difficulty of fast adaptation to solve downstream tasks.
These factors stem from the nature of RL and are inevitably necessary to be considered.

\begin{figure}
    \centering
    \includegraphics[width=\linewidth]{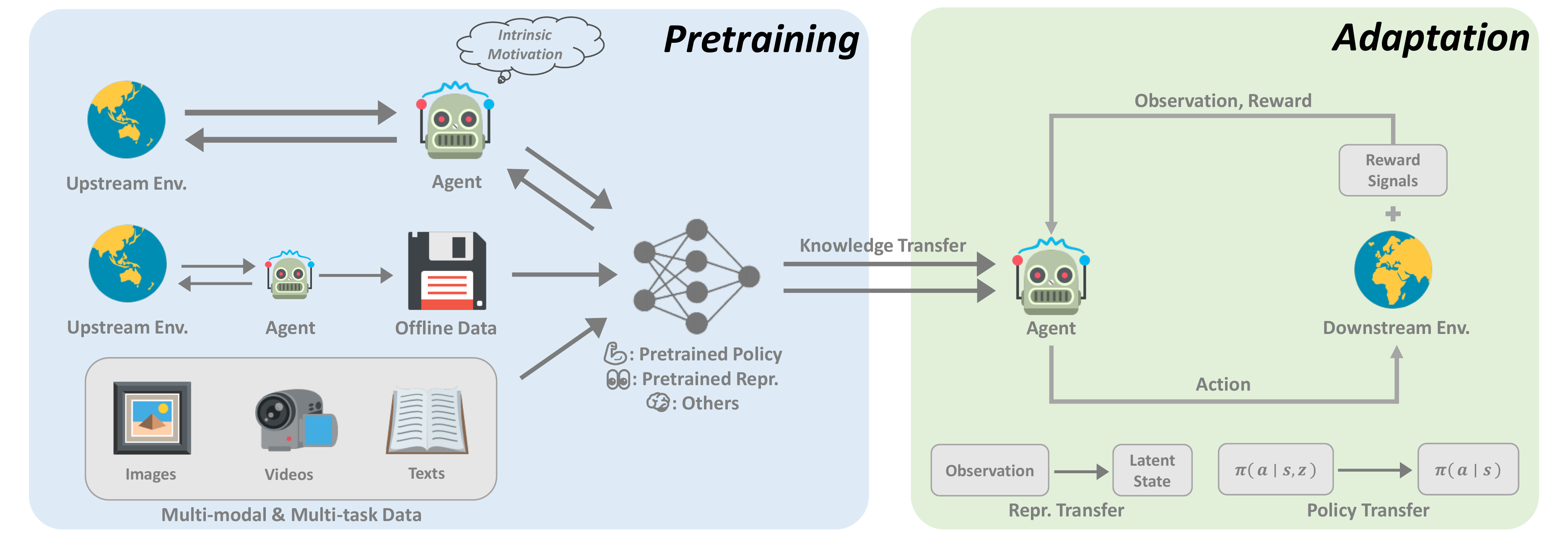}
    \caption{An illustrating example of the RL pretraining pipeline.}
    \label{fig:illustration}
\end{figure}

This survey aims to present a bird’s-eye view of current research on \textit{Pretraining in Deep RL}.
% , provide a taxonomy of these methods, discuss each sub-field thoroughly, and bring attention to open problems and prospects.
Principled pretraining in RL 
has a variety of potential benefits.
First of all, the substantial computational cost incurred by RL training remains a hurdle for industrial applications.
For example, replicating the results of AlphaStar~\shortcite{vinyals2019grandmaster} approximately costs millions of dollars~\shortcite{agarwal2022beyond}.
Pretraining can ameliorate this issue, either with pretrained world models~\shortcite{sekar2020planning} or pretrained representations~\shortcite{schwarzer2021pretraining}, by enabling quick adaptation to solve tasks in zero or few-shot manner.
Besides, RL is notoriously task- and domain-specific.
It has already been shown that pretraining with massive task-agnostic data can enhance these kinds of generalizations~\shortcite{lee2022multi}.
Finally, we believe that pretraining with proper architectures can unlock the power of scaling laws~\shortcite{kaplan2020scaling}, as shown by recent success in games~\shortcite{schwarzer2021pretraining,lee2022multi}.
By scaling up general-purpose models with increased computation, we are able to further achieve superhuman results, as taught in the ``bitter lesson''~\cite{sutton2019bitter}.

Pretraining in deep RL has undergone several breakthroughs in recent years.
\textit{Naive pretraining with expert demonstrations}, using supervised learning to predict the actions taken by experts, has been exhibited with the famed AlphaGo~\shortcite{silver2016mastering}.
To pursue large-scale pretraining with less supervision, the field of \textit{unsupervised RL} has been growing rapidly in recent years~\shortcite{burda2018largescale,laskin2021urlb}, which allows the agent to learn from interacting with the environment in the absence of reward signals.
In accordance with recent advances in offline RL~\cite{levine2020offline}, researchers further consider how to leverage unlabeled and sub-optimal offline data for pretraining~\shortcite{stooke2021decoupling,schwarzer2021pretraining}, which we term \textit{offline pretraining}.
The offline paradigm with task-irrelevant data further paves the way towards \textit{generalist pretraining}, where diverse datasets from different tasks and modalities as well as general-purpose models with great scaling properties are combined to build generalist models~\shortcite{reed2022generalist,lee2022multi}.
% Table~\ref{tab:setting} compares these settings from various perspectives.

% \input{tab/setting}

Pretraining has the potential to play a big role for RL and this survey could serve as a starting point for those interested in this direction.
In this paper, we seek to provide a systematic review of existing works in pretraining for deep reinforcement learning.
To the best of our knowledge, it is one of the pioneering efforts to systematically study pretraining in deep RL.

Following the development trend of pretraining in RL, we organize the paper as follows.
After going through the preliminaries of reinforcement learning and pretraining (Section~\ref{sec:pretrain}), we start with online pretraining in which an agent learns from interacting with the environment without reward signals (Section~\ref{sec:online}).
And then, we consider offline pretraining, the scenario where unlabeled training data is collected once with any policy (Section~\ref{sec:offline}).
In Section~\ref{sec:generalist}, we discuss recent advances in developing generalist agents for a variety of orthogonal tasks. 
We further discuss how to adapt to downstream RL tasks (Section~\ref{sec:downstream}).
% Some theoretical backends to RL pretraining research will be covered in Section~\ref{sec:theory}. 
Finally, we conclude this survey together with a few prospects (Section~\ref{sec:conclude}).

\section{Preliminaries}\label{sec:pretrain}

\begin{table}[t]
    \centering
    \begin{tabular}{ll}
        \thickhline
        \textbf{Notation} & \textbf{Description}\\
        \hline
        $\mathcal{M}$ & Markov decision process\\
        $\mathcal{S}$ & State space\\
        $\mathcal{A}$ & Action space\\
        $\mathcal{T}$ & Transition function\\ $\rho_{0}$ & Initial state distribution\\ $r$ & Reward function\\
        $\gamma$ & Discount factor\\
        $\mathcal{D}$ & Offline dataset\\
        $\tau$ & Trajectory\\
        $Q$ & Q function\\
        $J$ & Expected total discounted reward function\\
        $\theta$ & Neural network parameters\\
        $\phi$ & Feature encoder\\
        $z$ & Skill latent vector\\
        $\mathcal{Z}$ & Skill latent space\\
        $H$ & Entropy\\
        $I$ & Mutual information\\
        \thickhline
    \end{tabular}
    \caption{Notations used in the survey.}
    \label{tab:notation}
\end{table}

\subsection{Reinforcement learning}
Reinforcement learning considers the problem of finding a policy that interacts with the environment under uncertainty to maximize its collected reward.
Mathematically, this problem can be formulated via a Markov Decision Process (MDP) defined by tuple
($\mathcal{S}$, $\mathcal{A}$, $\mathcal{T}$, $\rho_{0}$, $r$, $\gamma$), with a state space $\mathcal{S}$, an action space $\mathcal{A}$, a state transition distribution $\mathcal{T}: \mathcal{S} \times \mathcal{A} \times \mathcal{S} \rightarrow[0,1]$, an initial state distribution $\rho_{0}: \mathcal{S} \rightarrow [0,1]$, a reward function $r: \mathcal{S} \times \mathcal{A} \rightarrow \mathbb{R}$, and a discount factor $\gamma \in (0,1)$.
The objective is to find such a policy $\policy$ parameterized by $\theta$ that maximizes

\begin{equation*}
    J(\pi_\theta)=\mathbb{E}_{\pi_\theta, \mathcal{T}, \rho_0}\left[\sum_{t=0}^{\infty} \gamma^t r\left(s_t, a_t\right)\right],
\end{equation*}
known as the discounted returns.
The notation used in the paper is summarized in Table~\ref{tab:notation}.
% We consider each MDP instance as a \textit{task}.
% MDPs with the same state space, action space, and transition function are considered to be of the same \textit{environment}.

% \zhihui{A primer on common algorithms.}
% \zhihui{VISR formulation.}

\subsection{Pretraining}

Pretraining aims at obtaining transferable knowledge from large-scale training data to facilitate downstream tasks.
In the context of RL,
\textit{transferable knowledge} typically includes good representations that facilitate the agent to perceive the world (i.e., a better state space) and reusable skills from which the agent can quickly build complex behaviors given task descriptions (i.e., a better action space).
\textit{Training data} can be one bottleneck for effective RL pretraining.
Unlike what we have witnessed in fields like computer vision and natural language processing where a wealth of unlabeled data can be collected with minimal supervision, RL usually requires highly task-specific reward design, which hinders scaling up pretraining for large-scale applications.

Therefore, the focus of this survey is \textit{unsupervised pretraining}, in which task-specific rewards are unavailable during pretraining but it is still allowed to learn from online interaction, unlabeled logged data, or task-irrelevant data from other modalities.
We omit supervised pretraining given that with task-specific rewards this scenario roughly degenerates to existing RL settings~\shortcite{levine2020offline}.
Figure~\ref{fig:illustration} demonstrates an overview of the pretraining and adaptation process.

The objective is to acquire useful prior knowledge in various forms like good visual representations, exploratory policies $\policy$, latent-conditioned policies $\cpolicy$, or simply logged datasets.
Depending on what data is available during pretraining, it requires different considerations to obtain useful knowledge (Section~\ref{sec:online}-\ref{sec:generalist}) and adapt it accordingly to downstream tasks (Section~\ref{sec:downstream}).

\begin{table*}[t]
    \centering
    \scriptsize
    \begin{tabular}{lllc}
        \thickhline
        \textbf{Type} & \textbf{Algorithm} & \textbf{Intrinsic Reward} & \textbf{Visual}\\
        \hline
        \multirow{4}{8em}{Curiosity-driven Exploration} & ICM~\shortcite{pathak2017curiosity} & $r_{t} \propto \left\|f\left(\phi\left(s_{t}\right), a_t\right)-\phi\left(s_{t+1}\right)\right\|^2$ & \cmark \\
        & RND~\shortcite{burda2018exploration} & $r_{t} \propto \left\|f\left(\phi\left(s_{t}\right), a_t\right)-\phi\left(s_{t+1}\right)\right\|^2$ & \cmark \\
        & Disagreement~\shortcite{pathak2019self} & $r_t \propto \operatorname{Var} \left(f\left(\phi\left(s_{t}\right), a_t\right)\right)$ & \cmark \\
        & Plan2Explore~\shortcite{sekar2020planning} & $r_t \propto \operatorname{Var} \left(f\left(\phi\left(s_{t}\right), a_t\right)\right)$ & \cmark \\
        \hline
        \multirow{13}{8em}{Skill Discovery}
        & VIC~\shortcite{gregor2016variational} & $r \propto \log q\left(z \mid \phi(s_H)\right)-\log p(z)$ & \cmark \\
        & VALOR~\shortcite{achiam2018variational} & $r \propto \log q\left(z \mid s_{1: H}\right)-\log p(z)$ & \xmark \\
        & DIAYN~\shortcite{eysenbach2018diversity} & $r_t \propto \log q\left(z \mid s_t\right)-\log p(z)$ & \xmark \\
        & VISR~\shortcite{Hansen2020Fast} & $r_t \propto \log q\left(z \mid \phi(s_t)\right)-\log p(z)$ & \cmark \\
        & DADS~\shortcite{Sharma2020Dynamics-Aware}  & $r_t \propto \log q\left(s_{t+1} \mid s_t, z\right)-\log q\left(s_{t+1} \mid s_t\right)$ & \xmark\\
        & EDL~\shortcite{campos2020explore} & $r_t \propto \log q\left(s_t \mid z\right)$ & \xmark \\
        & APS~\shortcite{liu2021aps} & $r_t \propto \log q\left(s_t \mid z\right)+\sum_{i \in \mathcal{I}_{\text{random}}} \log \left\|\phi(s_t)-h_i\right\|$ & \cmark \\
        % & HSD-3~\shortcite{gehring2021hierarchical} & $r=\left\|\omega^F\left(s^g\right)-g\right\|_2-\left\|\omega^F\left(s^{g^{\prime}}\right)-g\right\|_2$ & \xmark \\
        & HIDIO~\shortcite{zhang2021hierarchical} & $r_t \propto \log q(z \mid a_{t-k+1:t}, s_{t-k:t})$ & \xmark \\
        % & IBOL~\shortcite{kim2021unsupervised} & \xmark \\
        & UPSIDE~\shortcite{kamienny2022direct} & $r_t \propto \log q(z \mid s_t)-\log p(z)$ & \xmark \\
        & LSD~\shortcite{park2022lipschitzconstrained} & $r_t \propto \left(\phi\left(s_{t+1}\right)-\phi\left(s_t\right)\right)^{\top} z$ & \xmark \\
        \hline
        \multirow{6}{8em}{Data Coverage Maximization} & CBB~\shortcite{bellemare2016unifying} & $r_t \propto \hat{N}(s_t)^{-\frac{1}{2}}$ & \xmark \\
        & MaxEnt~\shortcite{hazan2019provably} & $r_t \propto \nabla R\left(\hat{d}_{\pi_{t}}\right)$ & \xmark \\
        & SMM~\shortcite{lee2019efficient} & $r_t \propto \log \hat{p}(s_t)-\log p_\pi(s_t)$ & \xmark \\
        & APT~\shortcite{liu2021behavior} & $r_t \propto \sum_{i \in \mathcal{I}_{\text{random}}} \log \left\|\phi(s_t)-h_i\right\|$ & \cmark \\
        & Proto-RL~\shortcite{yarats2021reinforcement} & $r_t \propto \sum_{i \in \mathcal{I}_{\text{prototype}}} \log \left\|\phi(s_t)-h_i\right\|$ & \cmark \\
        & RE3~\shortcite{seo2021state} & $r_t \propto \log \left(\left\|\phi(s_t) - \operatorname{KNN}\left(\phi(s_t)\right)\right\|+1\right)$& \cmark \\
        
        \thickhline
        
        % RND~\shortcite{burda2018exploration} & \\
        % Novelty~\shortcite{tao2020novelty} & \\
        % NGU~\shortcite{Badia2020Never} & & pseudo-count & \cmark \\
    \end{tabular}
    \caption{
    Categorization of representative online pretraining approaches.\
    % $\operatorname{Var}(\cdot)$ denotes the empirical variance over ensembles.
    % $R(\cdot)$ denotes a reward functional.
    % $\mathcal{I}_{\text{random}}$ and $\mathcal{I}_{\text{prototype}}$ denote 
    % \zhihui{
    % Some are borrowed from APT paper.
    % Will be updated soon.
    % Discussion about different rewards: \shortcite{park2022lipschitzconstrained}.
    % }
    }
    \label{tab:online}
\end{table*}

% \section{Naive Pretraining}
\section{Online Pretraining}\label{sec:online}

Most of the previous successes in RL have been achieved given dense and well-designed reward functions.
Despite its primacy in providing excel performances for a specific task, the traditional RL paradigm faces two critical challenges when scaling it up to large-scale pretraining.
Firstly, it is notoriously easy for an RL agent to overfit~\shortcite{zhang2018dissection}.
As a result, a pretrained agent trained with sophisticated task rewards can hardly generalize to unseen task specifications.
Furthermore, it remains a practical challenge to design reward functions which is usually costly and requires expert knowledge.

Online pretraining without these reward signals can potentially be a good solution to learning generic skills and eliminate the supervision requirement.
Online pretraining aims at acquiring prior knowledge by interacting with the environment in the absence of human supervision.
During the pretraining phase, the agent is allowed to interact with the environment for a long period without access to extrinsic rewards. 
When the environment is accessible, playing with it facilitates skill learning that will be useful later when a task is assigned to the agent.
This solution, also known as \textit{unsupervised RL}, has been actively studied in recent years~\shortcite{burda2018largescale,srinivas2021unsupervised}.

To encourage the agent to build its own knowledge without any supervision, we need principled mechanisms to provide the agent with intrinsic drives.
Psychologists found that babies can discover both the tasks to be learned and the solution to those tasks through interacting with the environment~\shortcite{smith2005development}.
With experiences accumulated, they are capable of more difficult tasks later on.
This motivates a wealth of research that studies how to build self-taught agents with \textit{intrinsic rewards}~\shortcite{schmidhuber1991curious,chentanez2004intrinsically,oudeyer2007intrinsic}.
Intrinsic rewards, in contrast to task-specifying extrinsic rewards, refer to general learning signals that encourage the agent either to collect diverse experiences or to develop useful skills.
It has been shown that pretraining an agent with intrinsic rewards and standard RL algorithms can lead to fast adaptation once the downstream task is given~\shortcite{laskin2021urlb}.

% Online pretraining with unsupervised RL is closely related to exploration.
% In spite of their similarity in using auxiliary incentives to take exploratory actions, they are usually built upon different motivations.
% Exploration techniques aim at facilitating policy learning when the extrinsic rewards are sparse.
% Online pretraining, on the other hand, requires learning useful prior knowledge for downstream tasks in the absence of extrinsic rewards.

Based on how to design intrinsic rewards, we classify existing approaches of unsupervised RL into three categories\footnote{This taxonomy of unsupervised RL was originally proposed by \shortciteA{srinivas2021unsupervised}.}: curiosity-driven exploration, skill discovery, and maximal data coverage.
Table~\ref{tab:online} presents a categorization of representative online pretraining algorithms together with their used intrinsic rewards.

% Model-based methods can learn world models in an unsupervised manner.
% However, it is essential to guarantee data diversity for accurate prediction.

\subsection{Curiosity-driven Exploration}\label{sec:curiosity}

\begin{figure}
    \centering
    \includegraphics[width=0.5\linewidth]{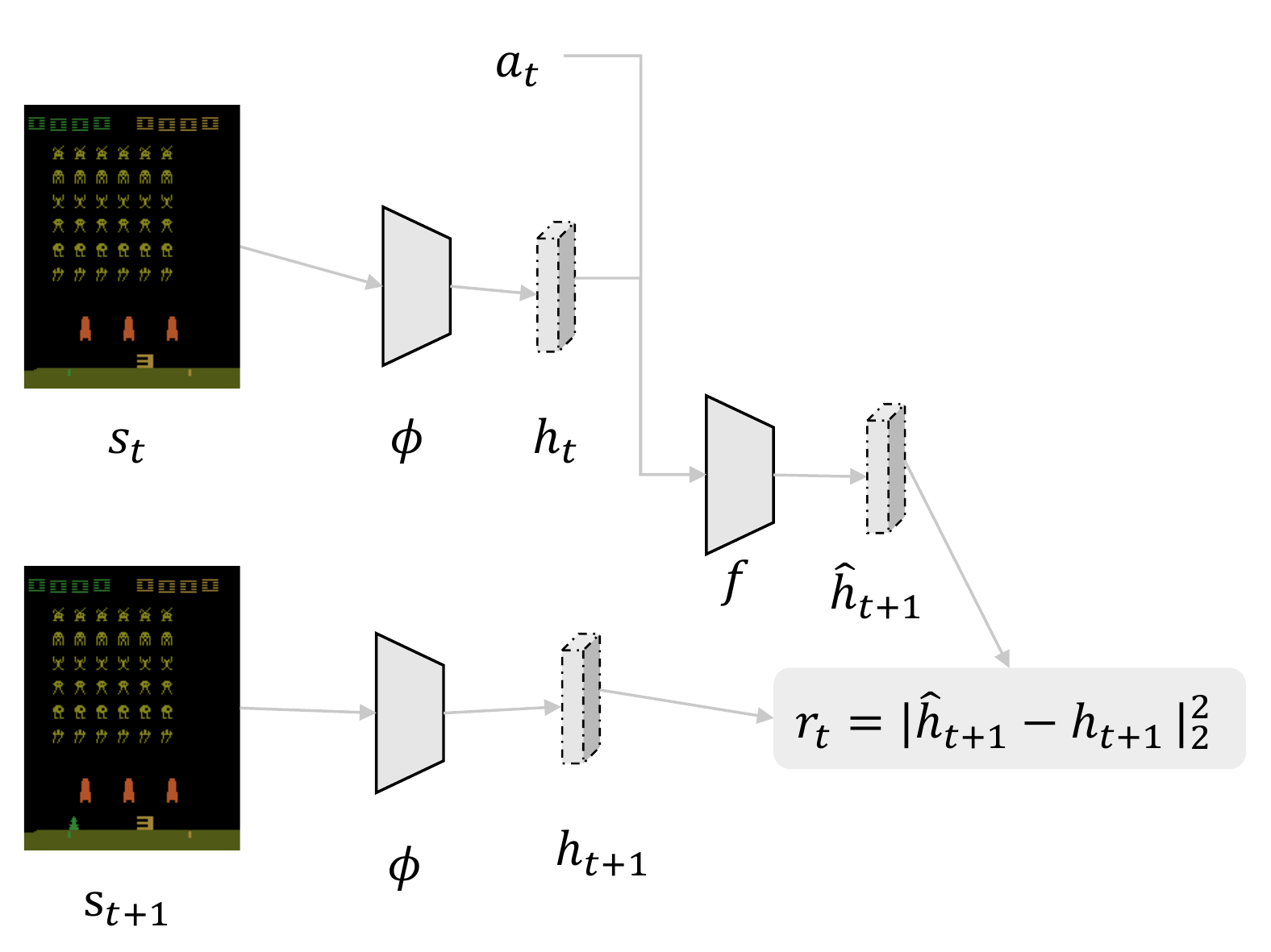}
    \caption{The process of computing intrinsic rewards using curiosity-driven exploration approaches.}
    \label{fig:curiosity}
\end{figure}

In the psychology of motivation, curiosity represents motivation to reduce uncertainty about the world~\shortcite{silvia2012curiosity}.
Inspired by this line of psychological theory, similar ideas have been studied to build curiosity-driven approaches for online pretraining.
Curiosity-driven approaches seek to explore \textit{interesting} states that can possibly bring knowledge about the environment.
Intuitively, if the agent falls short of accurately predicting the environment, it gains knowledge by interacting and then reducing this part of the uncertainty.
The defining characteristic of a curiosity-driven agent is how to compute the degree of curiosity to these interesting states, which directly serves as the intrinsic reward for learning.
A concrete example is ICM~\shortcite{pathak2017curiosity}, which applies the intrinsic reward proportional to the prediction error as shown in Figure~\ref{fig:curiosity}:
\begin{equation*}
    r_{t} \propto \left\|f\left(\phi\left(s_{t}\right), a_t\right)-\phi\left(s_{t+1}\right)\right\|_2^2,
\end{equation*}
where $f$ and $\phi$ represent the learned forward dynamics model and feature encoder, respectively.
% The forward dynamics model is trained by regression, whereas the feature encoder is trained with inverse dynamics modeling.

To measure curiosity, a broad class of approaches~\shortcite{pathak2017curiosity,haber2018learning} leverages this kind of learned dynamics models to predict future states in an auxiliary feature space.
there are mainly two kinds of estimation: prediction error and prediction uncertainty.
Despite that these dynamics-based approaches perform quite well across common scenarios, they usually suffer from action-dependent noisy TVs~\shortcite{burda2018largescale}, which will be discussed later in Section~\ref{sec:curiosity_challenge}.
This deficiency encourages the following work to design dynamics-free curiosity estimation~\shortcite{burda2018exploration} and more sophisticated uncertainty estimation methods~\shortcite{pathak2019self,sekar2020planning}.

Another important design choice is associated with the feature encoder $\phi$, especially for high-dimensional observations.
A proper feature encoder can make the
prediction task more tractable and filter out irrelevant aspects so that the agent can only focus on the informative ones.
Early studies~\shortcite{stadie2015incentivizing,burda2018largescale} leverage auto-encoding embeddings to recover the original high-dimensional inputs, but the induced feature space is usually too informative about irrelevant details and hence susceptible to noise.
To address this issue, \shortciteA{pathak2017curiosity} utilize an inverse dynamics model for feature encoding to make sure that the agent is unaffected by nuisance factors in the environment.
The proposed ICM shows impressive zero-shot performance in playing video games.
\shortciteA{burda2018exploration} further relax the design burden by simply replacing the feature model with a fixed randomly initialized neural network, which is proven effective by a following large-scale empirical study~\shortcite{burda2018largescale}.
Despite that random feature encoders are sufficient for good performance at training, learned features (e.g., based on inverse dynamics) generalize better~\shortcite{burda2018largescale}.
Inspired by recent advances in representation learning, \shortciteA{du2021curious} directly link curiosity and representation learning loss by formulating a minimax game between a generic representation learning algorithm and a reinforcement learning policy.

\subsubsection{Challenges \& Future Directions}\label{sec:curiosity_challenge}
This kind of approach has several deficiencies.
One of the most important issues is how to distinguish \textit{epistemic} and \textit{aleatoric} uncertainty.
Epistemic uncertainty refers to uncertainty caused
by a lack of knowledge.
Aleatoric uncertainty, in contrast, refers to the variability in the outcome due to  inherently random effects.
A concrete phenomenon in RL is the noisy TV problem~\shortcite{mavor2022stay}, which refers to the cases where the agent gets trapped by its curiosity in highly stochastic environments.
To mitigate this issue, some work attempts to use intrinsic rewards proportional to a reduction in uncertainty~\shortcite{houthooft2016vime,pathak2019self}.
However, tractable epistemic uncertainty estimation in high dimension remains challenging~\shortcite{hullermeier2021aleatoric} due to its sensitivity to imperfect data.

% \zhihui{uncertainty => reduction in uncertainty}.
% Reinforcement Learning with Prediction-Based Rewards. OpenAI Blog.

% Instead of minimizing prediction error, another line of methods focus on prediction uncertainty~\shortcite{houthooft2016vime,pathak2019self}.
% \shortcite{houthooft2016vime} learn probabilistic predictive models to measure uncertainty, and aim to maximize the reduction in uncertainty.
% \shortcite{pathak2019self} estimate the uncertainty with an ensemble of forward dynamics models.

Another issue with the above approaches is that they only receive retrospective signals after the agent has achieved epistemic uncertainty, which might cause inefficiency in exploration.
Based on this intuition, \shortciteA{sekar2020planning} design a model-based method that can prospectively look for uncertainty in the environment.

% \paragraph{Model-based Perspective}
% Model-based
% approaches can learn world models from unsupervised interaction [26]. However, the diversity of
% the training data will impact the accuracy of the model [57] and deploying this type of approach
% in visually complex domains like Atari remains an open problem [27].

\subsection{Skill Discovery}\label{sec:skill}

\begin{figure}
    \centering
    \includegraphics[width=0.5\linewidth]{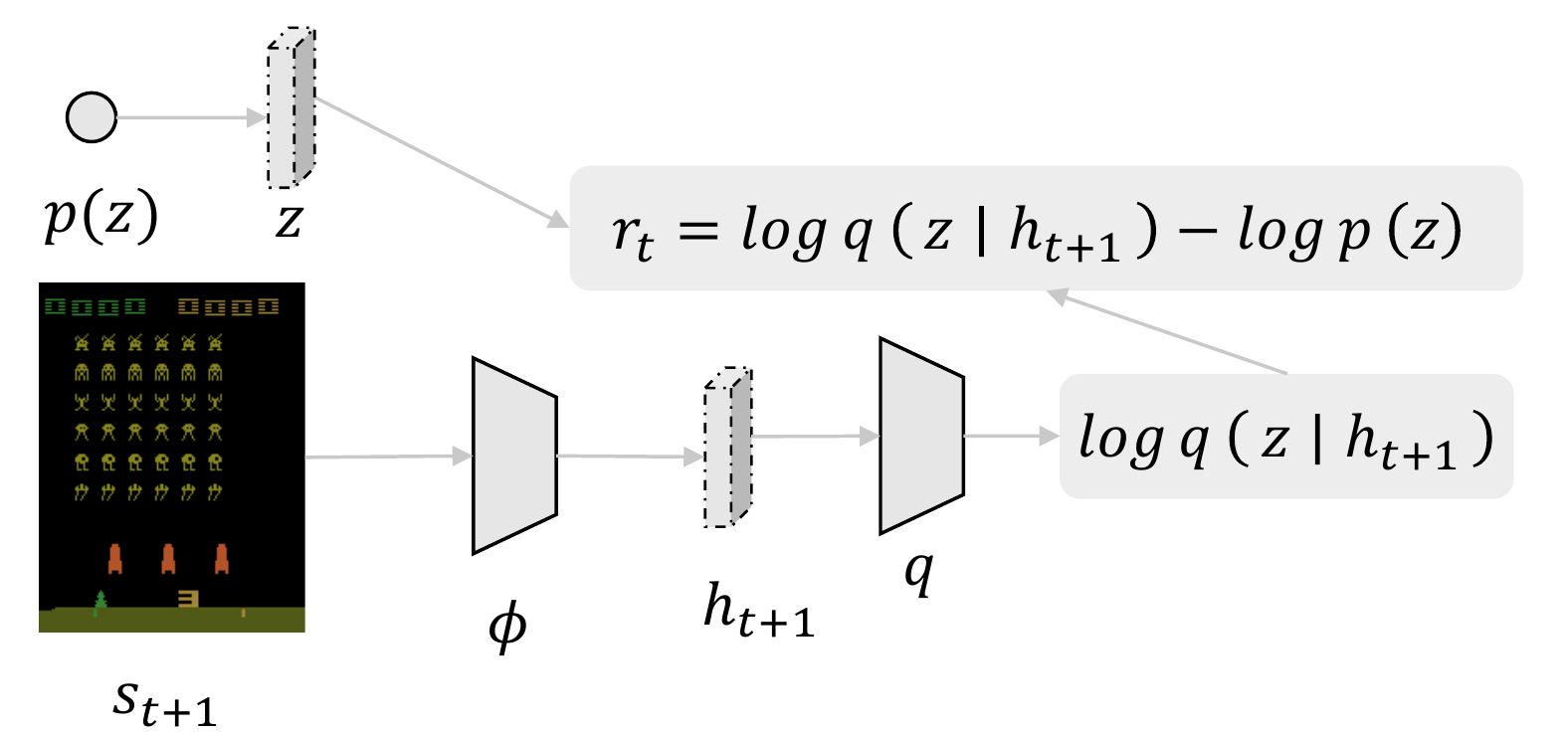}
    \caption{The process of computing intrinsic rewards using skill discovery approaches.}
    \label{fig:skill}
\end{figure}

Apart from curiosity-driven approaches that tackle unsupervised RL in a model-based perspective, one can also consider model-free learning of primitive skills\footnote{In this work, we use \textit{skill}, \textit{option}, and \textit{behavior prior} interchangeably.} that can be composed to solve downstream tasks.
This kind of approach is usually referred to as skill discovery approach.
The main intuition behind this is that the learned skill should control which states the agent visits, which can be seen as a notion of empowerment.

Generally speaking, the objective for skill discovery can be formalized as maximizing the mutual information (MI) between skill latent variable $z$ and state $s$:

\begin{equation}\label{eq:mi}
    I(s ; z) = H(z)-H(z \mid s) = H(s)-H(s \mid z),
\end{equation}
where we define \textit{skills} or \textit{options} as the policies conditioned on $z$.
There are two components for a skill discovery agent to determine: 1) a skill distribution $p(z)$; 2) a skill policy $\cpolicy$.
Before each episode, skill latent $z$ is sampled from distribution $p(z)$, followed by skill $\cpolicy$ to interact with the environment.
Learning skills that maximize MI is a challenging optimization problem, upon which a variety of approaches sharing the same spirit have been applied.

Among the existing MI-based skill discovery methods, the majority~\shortcite{gregor2016variational,eysenbach2018diversity,Hansen2020Fast} apply the former form of Equation~\ref{eq:mi} with the following variational lower bound~\shortcite{agakov2004algorithm}:
\begin{equation*}
    \begin{aligned}
        I(s ; z)
        &=\mathbb{E}_{s, z \sim p(s, z)}[\log p(z \mid s)]-\mathbb{E}_{z \sim p(z)}[\log p(z)] \\
        &\geq \mathbb{E}_{s, z \sim p(s, z)}\left[\log q(z \mid s)\right]-\mathbb{E}_{z \sim p(z)}[\log p(z)].
    \end{aligned}
\end{equation*}
In this case, a parametric model $q(z \mid s)$ is trained together with other variables to estimate the conditional distribution $p(z \mid s)$.
% Intuitively, the formulation simultaneously maximizes the entropy of $z$ to setup a broad class of target skills, and minimizes the conditional entropy of $z$ given $s$ to ensure the learned skill is predictable from the induced trajectory.
Maximizing $H(z)$ can be achieved by sampling $z$ from a learned distribution~\shortcite{gregor2016variational} or directly from a fixed uniform distribution~\shortcite{eysenbach2018diversity}.
As shown in Figure~\ref{fig:skill}, the intrinsic reward is given by $r_t=\log q\left(z \mid s_t\right)-\log p(z)$, upon which one can apply standard RL algorithms to learn skills.

% When $z$ is discrete, it suffices to fit a discriminator~\shortcite{gregor2016variational,eysenbach2018diversity} to minimize $H(z \mid s)$.
% For continuous $z$, techniques like contrastive approximation~\shortcite{warde-farley2018unsupervised} have also been studied to estimate $H(z \mid s)$.

% As shown in \shortcite{campos2020explore}, using the MI objective in Equation~\ref{eq:mi} with estimated distributions induced by the skill policy will incur larger rewards for known states than unexplored ones.

Another line of research~\shortcite{Sharma2020Dynamics-Aware,campos2020explore,liu2021aps,laskin2022cic} considers the latter form and similarly derives a lower bound:

\begin{equation*}
    \begin{aligned}
        I(s ; z)
        &=\mathbb{E}_{s, z \sim p(s, z)}[\log p(s \mid z)]-\mathbb{E}_{s \sim p(s)}[\log p(s)] \\
        &\geq \mathbb{E}_{s, z \sim p(s, z)}\left[\log q(s \mid z)\right]-\mathbb{E}_{s \sim p(s)}[\log p(s)].
    \end{aligned}
\end{equation*}
In this formulation, maximizing the state entropy $H(s)$ encourages exploration while minimizing the conditional entropy results in directed behaviors.
The difficulty lies in the density estimation of $s$, especially for high-dimensional state spaces.
A common practice is to maximize $H(s)$ via maximum entropy estimation~\shortcite{hazan2019provably,lee2019efficient,liu2021behavior}, which will be elaborated more in Section~\ref{sec:max-entropy}.
% When such a distribution is not available, it reduces to the problem of finding an exploration policy that maximizes the entropy of the induced distribution over the
% state space~\shortcite{hazan2019provably}.
% Minimizing $H(s \mid z)$.
% Note that unlike the situation for the former form, we now separate the skill policy from the sampling policy.
% \zhihui{trajectory-first}.
% \zhihui{Off-policy}.
% it becomes more tricky to optimize the objective.
% The difficulty lies in the density estimation of $s$, especially for high-dimensional state spaces.
% DADS~\shortcite{Sharma2020Dynamics-Aware} estimates $p(s)$ using $q_{\phi}(s \mid z)$ marginalized over sampled skill from a fixed prior $p(z)$.
% \zhihui{Connection to entropy maximization approaches.}
% EDL~\shortcite{campos2020explore} leverages maximum entropy exploration methods~\shortcite{hazan2019provably,lee2019efficient}.
% APS~\shortcite{liu2021aps} combines VISR~\shortcite{Hansen2020Fast} with a particle-based estimator as in \shortcite{liu2021behavior}.

Although different work uses slightly different approaches to optimize Equation~\ref{eq:mi}, it could be more important to decide other design factors when using skill discovery for online pretraining.
For instance, while most studies consider the episodic setting, some efforts have been made to extend MI-based skill discovery to non-episodic settings~\shortcite{xu2020continual,lu2021resetfree}.
It is also promising to consider a curriculum with an increasing number of skills to learn~\shortcite{achiam2018variational}.
Several other factors are also worth mentioning, such as whether skill latent $z$ is discrete~\shortcite{achiam2018variational} or continuous~\shortcite{warde-farley2018unsupervised}, whether the reward signals are dense~\shortcite{eysenbach2018diversity} or sparse~\shortcite{gregor2016variational}, and whether it works for image-based observations~\shortcite{warde-farley2018unsupervised}.

Skill discovery can be also reinterpreted as goal-conditioned policy learning, where $z$ as \textit{self-generated} and \textit{abstract} goal is sampled from a distribution instead of provided by the task.
One can also consider generating \textit{concrete} goals in a self-supervised manner~\shortcite{warde-farley2018unsupervised,pong2020skew} and derive a goal-conditioned reward function similarly from MI maximization.
DISCERN~\shortcite{warde-farley2018unsupervised} designs a non-parametric approach for goal sampling, maintaining a buffer of past observations that drifts as the agent collects new experiences.
Skew-Fit~\shortcite{pong2020skew} instead learns a maximum entropy goal distribution by increasing the entropy of a generative model in an iterative manner.
\shortciteA{choi2021variational} provide a more formal connection mainly from the perspective of goal-conditioned RL.
We refer the interested reader to \shortciteA{colas2022autotelic} for further discussion.

\subsubsection{Challenges \& Future Directions}

A major issue for MI-based skill discovery approaches is that the objective does not necessarily lead to strong state coverage as one can maximize $I(s; z)$ even with the smallest state variations~\shortcite{campos2020explore,park2022lipschitzconstrained}.
This lack of coverage can greatly limit their applicability to downstream tasks with complex environments~\shortcite{campos2020explore}.
To resolve this issue, some existing work explicitly uses $x$-$y$ coordinates as features to enforce state coverage induced by skills~\shortcite{eysenbach2018diversity,Sharma2020Dynamics-Aware}.
It is also explored to separate the learning process to first maximize $H(s)$ via maximum entropy estimation, followed by behavior learning~\shortcite{campos2020explore,liu2021aps}.

Moreover, it is empirically shown that skill discovery methods underperform other kinds of online pretraining methods, which may be due to restricted skill spaces~\shortcite{laskin2021urlb}.
This calls attention to dissecting what skills are learned.
In order to live up to their full potential, the discovered skills must strike a balance between generality (i.e., the applicability to a large variety of downstream tasks) and specificity (i.e., the quality of being useful to induce specific behaviors)~\shortcite{gehring2021hierarchical}.
It is also desired to avoid learning trivial skills~\shortcite{Sharma2020Dynamics-Aware,baumli2021relative}.

% \zhihui{
% Skill discovery: "hierarchical agents [21] or
% integrated within the universal successor features framework [2, 3, 8, 28].
% Their potential lack of
% coverage generally limits their applicability to complex downstream tasks"~\cite{campos2021beyond}.
% }

% \zhihui{
% HSD-3~\shortcite{gehring2021hierarchical} pretrains a hierarchy of skills with increased complexity to battle the specificity-generality trade-off.
% Other hierarchical approaches~\shortcite{zhang2021hierarchical,kim2021unsupervised}.
% }

\subsection{Data Coverage Maximization}\label{sec:max-entropy}

Previously we have discussed how to obtain knowledge or skills, measured by the agent's own capability, from unsupervised interaction.
Albeit indirectly related to the agent's ability, data diversity induced by online pretraining plays an essential role in deciding how well the agent obtains prior knowledge.
In the field of supervised learning, recent advances have shown that diverse data can enhance out-of-distribution generalization~\shortcite{hendrycks2019augmix} and robustness~\shortcite{hendrycks2020pretrained}.
Another supporting evidence is that most of the famed datasets are large and diverse~\shortcite{deng2009imagenet,wang2018glue}.
Motivated by the above considerations, it is desired to use data coverage maximization, usually measured by state visitation, as an objective to stimulate unsupervised learning.
% In the absence of prior knowledge on downstream tasks, data diversity plays an essential role in pretraining.
% To obtain diverse data for RL with online pretraining, it is feasible to maximize state visitation coverage induced by the policy.

% From a data-centric prospective, another straight-forward objective is to maximize state visitation coverage, either by pseudo-count estimation~\shortcite{bellemare2016unifying,tang2017exploration,ostrovski2017count} or entropy estimation.

\subsubsection{Count-based Exploration}

The first category of data coverage maximization is count-based exploration.
Count-based exploration methods directly use visit counts to guide the agent towards underexplored states~\shortcite{bellemare2016unifying,ostrovski2017count}.
For tabular MDPs, Model-based Interval Estimation with Exploration Bonuses~\shortcite{strehl2008analysis} provably turn state-action $N(s, a)$ counts into an exploration bonus reward:

\begin{equation}\label{eq:count}
    r_t \propto N(s_t, a_t)^{-1 / 2}.
\end{equation}

Built on Equation~\ref{eq:count}, a series of work has studied how to tractably generalize count bonuses to high-dimensional state spaces~\shortcite{bellemare2016unifying,ostrovski2017count,tang2017exploration}.
To approximate these counts in high dimensions, \shortciteA{bellemare2016unifying} introduce \textit{pseudo-counts} derived from a density model.
Specifically, the pseudo-count is defined as:

\begin{equation*}
    \hat{N}(s)=\frac{\rho_t(s)\left(1-\rho_t^{\prime}(s)\right)}{\rho_t^{\prime}(s)-\rho_t(s)},
\end{equation*}
where $\rho$ is a density model over state space $\mathcal{S}$, $\rho_t(s)$ is the density assigned to $s$ after training on a sequence of states $s_1, \ldots, s_t$, and $\rho_t^\prime(s)$ is the density of $s$ if $\rho$ were to be trained on $s$ one additional time.
Based on similar ideas, it has been shown that a better density model~\shortcite{ostrovski2017count} or a hash function~\shortcite{tang2017exploration,rashid2020optimistic} for computing state statistics can further improve performance.
Besides, a self-supervised inverse dynamics model as discussed in Section~\ref{sec:curiosity} can also be used to bias the count-based bonuses towards what the agent can control~\shortcite{Badia2020Never}.
% To relax the requirement of a learned density model, it has been studied to use hash functions~\shortcite{tang2017exploration} or random neural networks~\shortcite{burda2018exploration}

% Built on exploration methods that "turn state-action counts into a bonus reward"~\shortcite{strehl2008analysis}, several variants of intrinsic reward are designed~\shortcite{parisi2021interesting}:
% \begin{equation*}
%     r_t \propto \frac{1}{\sqrt{n\left(f\left(s_{t}\right)\right)}}.
% \end{equation*}
% NGU~\shortcite{Badia2020Never,campos2021beyond} computes pseudo-counts using $k$-nearest neighbors.
% "uses an intrinsic reward that combines per-episode and life-long novelty to explicitly encourage
% the agent to repeatedly visit all controllable states in the environment over an episode".
% "A self-supervised inverse dynamics model
% is used to train the embeddings of the nearest neighbour lookup, biasing the novelty
% signal towards what the agent can control".

% Another issue is \textit{derailment}, which refers to the situation 
% forget how to go back to previously visited states (\textit{detachment}), or try to return to a promising state (\textit{derailment}).

\subsubsection{Entropy Maximization}

To encourage novel state visitation, an alternative objective is to directly maximize the entropy of state visitation distribution $d_{\pi}$ induced by policy $\policy$:

\begin{equation*}
    \pi^{*} \in \underset{\pi \in \Pi}{\arg \max} H\left(d_{\pi}\right),
\end{equation*}
where $H(\cdot)$ can be Shannon entropy~\shortcite{hazan2019provably,lee2019efficient,seo2021state}, Rényi entropy~\shortcite{zhang2021exploration}, or geometry-aware entropy~\shortcite{guo2021geometric}.
The state distribution $d_{\pi}$ can either be a discounted distribution~\shortcite{hazan2019provably}, a marginal distribution~\shortcite{lee2019efficient}, or a stationary distribution~\shortcite{tarbouriech2019active}.

% "It has been argued that exploring the environment efficiently will
% serve as a proxy for developing such behaviors [38], and exploration bonuses have been shown to
% produce meaningful behavior in the absence of reward [49, 10]".~\shortcite{campos2021beyond}
% However, prediction error based methods and pseudo count based methods have no guarantee to converge to an exploratory policy.
% Asymptotically, each state will be visited infinitely many times and hence the exploration bonus goes to zero.

% From a data-centric perspective, we can alternatively optimize the state visitation so as to maximize coverage.
% The objective is to directly maximize the entropy of state visitation distribution $d_{\pi}$:

% \begin{equation*}
%     \pi^{*} \in \underset{\pi \in \Pi}{\arg \max} H\left(d^{\pi}\right),
% \end{equation*}
% where $H(\cdot)$ can be Shannon entropy~\shortcite{hazan2019provably,lee2019efficient,seo2021state} or other kinds of entropy measure~\shortcite{guo2021geometric,zhang2021exploration}.
% The state distribution $d_{\pi}$ can either be a discounted distribution~\shortcite{hazan2019provably}, a marginal distribution~\shortcite{lee2019efficient}, or a stationary distribution~\shortcite{tarbouriech2019active}.

Albeit compelling, the objective relies on maximizing state entropy, which is notoriously hard to estimate and optimize.
\shortciteA{hazan2019provably} contribute a provably efficient algorithm in the tabular setting using the conditional gradient method~\shortcite{frank1956algorithm} to avoid direct optimization.
\shortciteA{lee2019efficient} propose a similar approach that can be viewed from the perspective of state marginal matching between the state distribution and a given target distribution (e.g., a uniform distribution).
Both \shortciteA{hazan2019provably} and \shortciteA{lee2019efficient} propose to learn a mixture of policies that maximizes the induced state entropy in an iterative manner.
While impressive, these parametric approaches struggle to scale up to high dimensional spaces.
To address this issue, \shortciteA{mutti2020a} instead optimize a non-parametric, particle-based estimate of state distribution entropy~\shortcite{kontoyiannis1998nonparametric}, but restrict its use to state-based tasks.

For unsupervised online pretraining with visual observations, entropy maximization becomes more tricky as exploration is now inextricably intertwined with representation learning.
This leads to a \textit{chicken-and-egg} problem~\shortcite{yarats2021reinforcement,tam2022semantic}, where learning useful representations requires diverse data, while effective exploration can only be achieved with good representations.
Based on particle-based entropy estimators, several approaches successfully apply entropy maximization in image-based tasks with self-supervised representations learned by inverse dynamics prediction~\shortcite{seo2021state}, contrastive learning~\shortcite{liu2021behavior,yarats2021reinforcement}, or the information bottleneck~\shortcite{tao2020novelty}.

\subsubsection{Challenges \& Future Directions}

Although count-based approaches are shown effective for exploration, it has been shown in previous work~\shortcite{ecoffet2021first} that they usually suffer from \textit{detachment}, in which the agent loses track of interesting areas to explore, and \textit{derailment}, in which the exploratory mechanism prevents it from returning to previously visited states.
Count-based approaches also tend to be short-sighted, driving the agent to get stuck in local minima~\shortcite{burda2018exploration}.

When applying state entropy maximization approaches for pretraining, it is worth pointing out that many of them aim at maximizing the entropy of all states visited during the process, and hence the final policy is not necessarily exploratory~\shortcite{lee2019efficient}.
It has also been shown theoretically that the class of Markovian policies is insufficient for the maximum state entropy objective, while non-Markovian policies are essential to guarantee good exploration.
% there exist several limitations as shown in \shortcite{ecoffet2021first} (mainly for count-based approaches).
% Specifically, these algorithms stop trying to reach visited states that could possibly lead to the discovery of new areas (\textit{detachment} and \textit{derailment}), as each visitation lowers the intrinsic reward.

Instead of learning an exploratory policy, another line of research considers collecting unlabeled records as a prerequisite for offline RL~\shortcite{yarats2022dont,lambert2022challenges}, which is an interesting direction for understanding and  utilizing task-agnostic agents.

% It is shown that these consumable intrinsic rewards cause
% the problem of \textit{detachment}~\shortcite{ecoffet2021first}.
% Another issue is \textit{derailment}.
% Furthermore, maximizing the state entropy during training does not necessarily result in good exploratory agents.

% However, prediction error based methods and pseudo count based methods have no guarantee to converge to an exploratory policy.
% Asymptotically, each state will be visited infinitely many times and hence the exploration bonus goes to zero.

% It is also worth mentioning that designing intrinsic rewards is task-dependent.
% \zhihui{To improve} Exploration in games can be in line with the designers' wills, whereas for real-world tasks in which safety is of high priority, it can be strongly prohibited to act randomly.

\section{Offline Pretraining}\label{sec:offline}
\begin{table*}[t]
    \centering
    \scriptsize
    \begin{tabular}{lllcc}
        \thickhline
        \textbf{Type} & \textbf{Algorithm} & \textbf{Objective} & \textbf{Visual} & \textbf{Expert Data}\\
        \hline
        \multirow{6}{7em}{Skill Extraction} & SPiRL~\shortcite{pertsch2020accelerating} & Variational Auto-encoder & \cmark & \xmark\\
        & OPAL~\shortcite{ajay2021opal} & Variational Auto-encoder & \xmark & \cmark \\
        & Parrot~\shortcite{singh2021parrot} & Normalizing Flow  & \cmark & \cmark\\
        & SkiLD~\shortcite{pertsch2021demonstrationguided} & Variational Auto-encoder & \xmark & \xmark\\
        & TRIAL~\shortcite{yang2022trail} & Energy-based Model & \xmark & \cmark\\
        & FIST~\shortcite{hakhamaneshi2022hierarchical} & Variational Auto-encoder & \xmark & \cmark \\
        % & HeLMS~\shortcite{rao2022learning} & \cmark & \\
        \hline
        \multirow{5}{7em}{Representation Learning} & World Model~\shortcite{ha2018world} & Reconstruction & \cmark & \xmark\\
        & ST-DIM~\shortcite{anand2019unsupervised} & Forward Pixel Prediction & \cmark & \xmark\\\
        & ATC~\shortcite{stooke2021decoupling} & Forward Dynamics Modeling & \cmark & \cmark\\
        & SGI~\shortcite{schwarzer2021pretraining} & Forward Dynamics Modeling & \cmark & \xmark \\
        & Markov~\shortcite{allen2021learning} & Inverse Dynamics Modeling & \cmark & \xmark \\
        % & PBL~\shortcite{guo2020bootstrap} & \\
        % & DIM~\shortcite{mazoure2020deep} & \\
        \thickhline
    \end{tabular}
    \caption{
    Categorization of representative offline pretraining approaches.}
    \label{tab:offline}
\end{table*}

Despite its attractive effectiveness of learning without human supervision, online pretraining is still limited for large-scale applications.
Eventually, it is difficult to reconcile online interaction with the need to train on large and diverse datasets~\shortcite{levine2021understanding}.
To address this issue, it is desired to decouple data collection and pretraining and directly leverage historical data collected from other agents or humans.

A feasible solution is offline RL~\shortcite{lange2012batch,levine2020offline}, which has been gaining attention recently.
Offline RL aims to obtain a reward-maximizing policy purely from offline data.
A fundamental challenge of offline RL is the \textit{distributional shift}, which refers to the distribution discrepancy between training data and those seen during testing.
Existing offline RL approaches focus on how to address this challenge when using function approximation.
For instance, policy constraint approaches~\shortcite{kumar2019stabilizing,Siegel2020Keep} explicitly require the learned policy to avoid taking unseen actions in the dataset.
Value regularization methods~\shortcite{kumar2020conservative} alleviate the overestimation problem of value functions by fitting them to some forms of lower bounds.
However, it remains under-explored whether policies trained offline can generalize to new contexts unseen in the offline dataset~\shortcite{kirk2021survey}.

Another scenario is offline-to-online RL~\shortcite{nair2020accelerating,lu2021aw,lee2022offline,kostrikov2022offline}, where offline RL is used for pretraining, followed by online finetuning.
It has been shown in this scenario that offline RL can accelerate online RL~\shortcite{nair2020accelerating}.
However, both offline RL and offline-to-online RL require the offline experience to be annotated with rewards, which are challenging to provide for large real-world datasets~\shortcite{pertsch2020accelerating}.

% However, offline RL methods require reward signals.

A compelling alternative direction for leveraging offline data is to sidestep policy learning, but instead learn prior knowledge that is beneficial for downstream tasks in terms of convergence speed or final performances.
What is more intriguing, if our model were able to utilize data without human supervision, it could potentially benefit from web-scale data for decision-making.
We refer to this setting as offline pretraining, where  the agent can extract important information (e.g., good representations and behavior priors) from offline data.
% Reward signals provided when finetuning allows the agent to adapt to the task.
In Table~\ref{tab:offline}, we categorize existing offline pretraining approaches as well as summarize each approach's key properties.
% This is in line with previous success on supervised learning, where pretrained models are widely-used.

% Unlike Section~\ref{sec:online} that focuses on how to collect experiences in a unsupervised fashion, here we care about how to learn from offline data collected a prior.
% There is a blur barrier between the approaches used in Section~\ref{sec:online} and those discussed here.

\subsection{Skill Extraction}

Learning useful behaviors from offline data has a long history~\shortcite{pomerleau1988alvinn,argall2009survey}.
When the offline data comes from expert demonstrations, it is straightforward to pretrain policies via imitation learning~\shortcite{silver2016mastering,rajeswaran2017learning,gupta2019relay}, which is often used in real-world applications like robotic manipulation~\shortcite{zhu2018reinforcement,zhang2018deep} and self-driving~\shortcite{codevilla2019exploring}. 
However, imitation learning approaches often assume that the training data contains complete solutions.
They therefore usually fall short of obtaining good policies when demonstrations are collected from a series of sources.

An alternative solution is to learn useful behavior priors from offline data~\shortcite{lynch2020learning,shankar2020learning,chebotar2021actionable}, similar to what we have discussed in Section~\ref{sec:skill}.
Compared with its online counterpart, offline skill extraction assumes a fixed set of trajectories.
These approaches learn a spectrum of behavior policies conditioned on latent $z$, which provide a more compact action space for learning high-level policies that can quickly adapt to downstream tasks.
% "Recently, a number of works have explored the embedding of skills into a continuous skill space via stochastic latent variable models [11, 7, 30, 8, 9, 31, 10]."~\shortcite{pertsch2020accelerating}
Specifically, temporal skill extraction~\shortcite{yang2022trail} for few-shot imitation~\shortcite{ajay2021opal} and RL~\shortcite{ajay2021opal,pertsch2020accelerating,pertsch2021demonstrationguided} considers how to distill offline trajectories into primitive policies $\cpolicy$, where $z \in \mathcal{Z}$ denotes a skill latent learned via unsupervised learning.
By leveraging stochastic latent variable models, we aim at learning a skill latent
$z_{i} \in \mathcal{Z}$ for a sequence of state-action pairs $\left\{s_{t}, a_{t}, \ldots, s_{t+H-1}, a_{t+H-1}\right\}$, where $H$ is a fixed horizon or a variable one~\shortcite{kipf2019compile,Shankar2020Discovering}.
For example, \shortciteA{ajay2021opal} propose the following auto-encoding objective to learn primitive skills:

\begin{equation*}
    \begin{aligned}
    &\min _{\theta, \phi, \omega} J(\theta, \phi, \omega)=\mathbb{E}_{\tau \sim \mathcal{D}, z \sim q_\phi(z \mid \tau)}\left[-\sum_{t=0}^{H-1} \log \pi_\theta\left(a_t \mid s_t, z\right)\right] \\
    &\text { s.t. } \mathbb{E}_{\tau \sim \mathcal{D}}\left[\mathrm{D}_{\mathrm{KL}}\left(q_\phi(z \mid \tau) \| \rho_\omega\left(z \mid s_0\right)\right)\right] \leq \epsilon_{\mathrm{KL}},
    \end{aligned}
\end{equation*}
where $q_\phi(z \mid \tau)$ encodes the trajectory $\tau$ into skill latent $z$ and skill policy $\cpolicy$ serves as a decoder to translate skill latent $z$ into action sequences.
% A generative action decoder $p(a \mid z)$ that translates skill embedding $z$ into action sequences $a$ is learned, together with a skill prior $p(z \mid s)$.
To transfer skills into downstream tasks, it is feasible to learn a hierarchical policy that generates high-level behaviors with $\pi(z \mid s)$ trained on downstream tasks~\shortcite{pertsch2020accelerating}, which will be elaborated in Secition~\ref{sec:policy_transfer}.
% "By selecting skills from the behavioral prior, the RL algorithm is able to explore in a structured manner and can solve long-horizon navigation and manipulation tasks."~\shortcite{hakhamaneshi2022hierarchical}

% \zhihui{
% In contrast to online pretraining, it is now trajectory-first.
% "Behavioral priors learned through maximum likelihood latent variable models have been used for
% structured exploration in RL Singh et al. (2020), to solve complex long-horizon tasks from sparse rewards
% Pertsch et al. (2020), and regularize offline RL policies Wu et al. (2019); Peng et al. (2019); Nair et al. (2020).
% While impressive, RL with data-driven behavioral priors does not generalize to new tasks efficiently, often
% requiring millions of environment interactions to converge to an optimal policy for a new task"
% }

% SPiRL~\shortcite{pertsch2020accelerating} defines a skill as a sequence of $H$ consecutive actions $\boldsymbol{a}=\left\{a_{t}, \ldots, a_{t+H-1}\right\}$.
% The finetuning objective poses constraints "that the policy remains close to the learned skill prior, guiding
% exploration during RL“.

% \begin{equation*}
%     J(\theta)=\mathbb{E}_{\pi_{\theta}}\left[\sum_{t=0}^{T-1} r\left(s_{t}, z_{t}\right)-\alpha D_{\mathrm{KL}}\left(\pi_{\theta}\left(z_{t} \mid s_{t}\right), p_{\boldsymbol{a}}\left(z_{t} \mid s_{t}\right)\right)\right]
% \end{equation*}
Various latent variable models have been used for pretraining behavior priors.
For instance, variational auto-encoders~\shortcite{kingma2013auto} are widely considered~\shortcite{pertsch2020accelerating,lynch2020learning,ajay2021opal}.
Following work~\shortcite{singh2021parrot,florence2022implicit} also explores normalizing flow~\shortcite{dinh2017density} and energy-based models~\shortcite{lecun2006tutorial} to learn action priors.

% SPiRL~\shortcite{pertsch2020accelerating}, Play-LMP~\shortcite{lynch2020learning}, OPAL~\shortcite{ajay2021opal};
% Normalizing flow~\shortcite{dinh2017density}: PARROT~\shortcite{singh2021parrot}.
% Energy-based model: \shortcite{florence2022implicit}, which shows impressive competence on multi-modal data, outperforming common explicit (Mean
% Square Error, or Mixture Density) behavioral cloning policies.
% Transformer~\shortcite{shafiullah2022behavior}.

% "Behavioral priors learned through maximum likelihood latent variable models have been used for
% structured exploration in RL Singh et al. (2020), to solve complex long-horizon tasks from sparse rewards
% Pertsch et al. (2020), and regularize offline RL policies Wu et al. (2019); Peng et al. (2019); Nair et al. (2020).
% While impressive, RL with data-driven behavioral priors does not generalize to new tasks efficiently, often
% requiring millions of environment interactions to converge to an optimal policy for a new task"

% Connection to skill discovery in unsupervised RL: "In contrast, OPAL focuses on settings where a large dataset of diverse
% behaviors is provided but access to the environment is restricted. It leverages these static offline
% datasets to discover primitive skills with better state coverage and avoids the exploration issue of
% learning primitives from scratch"~\shortcite{ajay2021opal}.

% \zhihui{Trajectory-first approaches~\shortcite{kim2021unsupervised}, in contrast to latent-first approaches, can be done off-policy.}

The scenario of pretraining behavior priors also bears resemblance to few-shot imitation learning~\shortcite{dance2021conditioned,hakhamaneshi2022hierarchical}.
However, for few-shot imitation learning, it is often assumed that expert data is collected from a single behavior policy.
Furthermore, due to error accumulation~\shortcite{ross2011reduction}, few-shot imitation learning is often limited to short-horizon problems~\shortcite{hakhamaneshi2022hierarchical}.
In this regard, learning behavior priors from diverse and sub-optimal data appears to be a promising direction.

\subsubsection{Challenges \& Future Directions}

Despite its potential to extract useful primitive skills, it is still challenging to pretrain on highly sub-optimal offline data containing random actions~\shortcite{ajay2021opal}.
Besides, RL with learned skills does not usually generalize to downstream tasks efficiently, requiring millions of online interactions to converge~\shortcite{hakhamaneshi2022hierarchical}.
A possible solution is to combine with successor features~\shortcite{barreto2017successor,Hansen2020Fast} for fast task inference.
However, strategies that directly
use the pretrained policies for exploitation may result in sub-optimal solutions in such a scenario~\shortcite{campos2021beyond}.

% \zhihui{Trade-off between generality (wide applicability) and
% specificity (benefits for specific tasks)~\shortcite{gehring2021hierarchical}.}

% \zhihui{However, strategies that directly
% use the pre-trained policies for exploitation may result in sub-optimal solutions in such scenario [2]~\shortcite{campos2021beyond}.}

% Despite that pretraining skills can be beneficial, it has several limitations: 1) it requires powerful generative models; 2) the learned skills might lead to local minima when finetuning; 3) it entangles representation learning and policy learning.

\subsection{Representation Learning}\label{sec:pretrain_repr}

% \zhihui{
% This section needs reorganization (maybe following https://arxiv.org/pdf/2207.08229.pdf), and ``failures of prior approaches'' in https://openreview.net/pdf?id=RQLLzMCefQu
% }

\begin{table}[t]
    \centering
    \begin{tabular}{lcc}
        \thickhline
        \textbf{Type} & \textbf{Sufficiency} & \textbf{Compactness} \\
        \hline
        Reconstruction & $\bigstar \bigstar \bigstar$ & $\bigstar$\\
        Forward Pixel Prediction & $\bigstar \bigstar \bigstar$ & $\bigstar$\\
        Forward Dynamics Modeling & $\bigstar \bigstar$ & $\bigstar\bigstar$\\
        Inverse Dynamics Modeling & $\bigstar$ & $\bigstar \bigstar \bigstar$\\
        \thickhline
    \end{tabular}
    \caption{
    Comparison between different representation learning approaches.
    % \zhihui{Five main papers: ACL~\shortcite{yang2021representation}, ATC~\shortcite{stooke2021decoupling}, SGI~\shortcite{schwarzer2021pretraining}, Markov~\shortcite{allen2021learning}, HOMER~\shortcite{misra2020kinematic}.
    % Also consider EIRLI's Figure 3.
    % }
    }
    \label{tab:representation}
\end{table}

While pretraining behavior priors focus on reducing the complexity of the action space, there exists another line of work that aims to pretrain good state representations from offline data to promote transfer.
If the agent effectively reduces the representation gap between the learned state representations and the ground-truth endogenous states, it can better focus on factors that are essential for control.
Table~\ref{tab:representation} compares different kinds of representation learning objectives in terms of sufficiency (i.e., whether the representations contain sufficient state information) and compactness (i.e., whether the representations discard irrelevant information).

Learning good state representations for RL is a mature research area with a range of tools~\shortcite{nachum2018nearoptimal,oord2018representation,zhang2022efficient}.
Traditionally, the problem is formulated to group states into clusters based on certain properties~\shortcite{li2006towards}.
Existing representation learning approaches generally propose some predictive properties that the desired representations have, with regard to states, actions, and rewards across different time-steps.
One of the most representative concepts is bisimulation~\shortcite{larsen1991bisimulation,givan2003equivalence}, which originally requires two equivalent states to have the same reward and equivalent distributions over the next bisimilar
states.
The objective turns out to be very restrictive and is further relaxed by following work~\shortcite{ferns2004metrics,castro2010using} with a defined pseudo-metric space to measure behavioral similarity.
Despite their recent advances~\shortcite{gelada2019deepmdp,zhang2021learning,agarwal2021contrastive} in effective representation learning using deep neural networks, bisimulation methods fail to provide good abstraction when the rewards are sparse or even absent.
In this case, solely relying on a forward model can lead to representation collapse~\shortcite{allen2021learning}.

To alleviate representation collapse, one can instead set the targets to pixel observations.
This includes reconstruction-based approaches~\shortcite{lange2010deep,ha2018world} and those based on pixel prediction~\shortcite{watter2015embed,kaiser2019model}.
Reconstruction-based approaches typically train an auto-encoder on image observations to learn a low-dimensional representation, using which a policy is learned subsequently.
Approaches based on pixel prediction force the representations to contain sufficient information about future pixel observations.
Despite these learned representations preserving sufficient information about the observation, it lacks compactness and does not guarantee to capture of useful information for the control task.
% Despite its advantages, predicting ground states provides a poor abstraction because task-irrelevant information is strictly preserved.

Instead of predicting the future, it is also beneficial to model the inverse dynamics of the system~\shortcite{christiano2016transfer,pathak2017curiosity}.
Inverse dynamics modeling learns a representation that is predictive of the action taken between a pair of consecutive states.
It has been shown that the learned representation can filter out all uncontrollable aspects of the observations~\shortcite{efroni2021provable}.
However, it can also wrongly ignore controllable information and cause over-abstraction over the state space~\shortcite{rakelly2021mutual,efroni2021provable}.

With the rise of self-supervised learning developed for CV and NLP, a natural direction is to adapt these task-agnostic techniques to RL.
For instance, a large body of works has explored contrastive learning~\shortcite{gutmann2010noise} as an effective framework to learn good representations~\shortcite{nachum2018nearoptimal,laskin2020curl,mazoure2020deep,zhan2020framework}.
Contrastive learning typically uses the InfoNCE loss~\shortcite{poole2019variational} to maximize mutual information between two variables:
\begin{equation*}
    \mathcal{L}_{\text {InfoNCE }}=\mathbb{E}\left[\log \frac{\exp \left(f\left(x_i, y_i\right)\right)}{\frac{1}{K} \sum_{j=1}^K \exp \left(f\left(x_i, y_j\right)\right)}\right],
\end{equation*}
where $f$ is a bilinear function $f\left(x_i, y_i\right)=\phi(x_i)^\top W \phi(y_i)$ with learned parameter $W \in \mathbb{R}^{n \times n}$ and $K$ is the number of negative samples.
These approaches usually incorporate temporal information, aiming to distinguish between sequential and non-sequential states~\shortcite{anand2019unsupervised,stooke2021decoupling}.
Following works~\shortcite{schwarzer2020data,schwarzer2021pretraining} further consider bootstrapped latent representations~\shortcite{grill2020bootstrap} that get rid of negative samples. 
% Bootstrapping~\shortcite{schwarzer2020data,schwarzer2021pretraining}.
% Some propose temporal contrastive learning objectives.
% \shortcite{zhan2020framework} use collected demonstrations to pretrain a visual encoder with a contrastive loss.

% \paragraph{Forward Dynamics Models}
% Predicting the future~\shortcite{guo2020bootstrap,schwarzer2020data}

Aside from the above representation learning objectives, some other work considers imposing Lipschitz smoothness~\shortcite{gelada2019deepmdp,zhang2021learning}, kinematic inseparability~\shortcite{misra2020kinematic}, or the Markov property~\shortcite{allen2021learning}.
It has also been shown that a combined objective can also lead to better performance~\shortcite{schwarzer2021pretraining}.
% mutual information maximization~\shortcite{anand2019unsupervised,lee2020predictive,rakelly2021mutual}, 

% \paragraph{Other Perspectives}
% Goal-conditioned (hindsight?)~\shortcite{chebotar2021actionable}; Their combination~\shortcite{schwarzer2021pretraining}; Kinematic Inseparability: "The idea behind KI abstractions is that unless two states can be distinguished from each other—by
% either their backward or forward dynamics—they ought to be treated as the same abstract state."; Lipschitz Smoothness: DeepMDP~\shortcite{gelada2019deepmdp} and DBC~\shortcite{zhang2021learning}; Mutual information maximization (ST-DIM~\shortcite{anand2019unsupervised} (across space and time), \shortcite{lee2020predictive,rakelly2021mutual}); Markov: \shortcite{allen2021learning}.

\subsubsection{Challenges \& Future Directions}

While unsupervised representations have been shown to bring significant improvements to downstream tasks, the absence of reward signals typically leads the pretrained encoder to focus on task-irrelevant features instead of task-relevant ones in visually complex environments~\shortcite{yamada2022task}.
To alleviate this issue, one might incorporate additional inductive bias~\shortcite{janny2022filtered} or labeled data that are cheaper to obtain.
We will discuss the latter solution in Section~\ref{sec:generalist}.

Another challenge for unsupervised representation learning is how to measure its effectiveness without access to downstream tasks.
Such evaluation is beneficial because it can provide a proxy metric to predict performance and promote a deeper understanding of the semantic meanings of pretrained representations.
To achieve this, it is desired to analyze these representations with probing techniques and determine which properties they encode.
Although previous work has made efforts in this direction~\shortcite{zhang2022light}, it remains unclear what properties are most indispensable for pretrained representations. 
% These representations can hardly capture good saliency maps~\shortcite{yamada2022task}.

% "However, follow-up results from RAD (Laskin et al., 2020)
% suggest that most of the benefits of CURL come from image augmentation, not its contrastive loss".

% Recent studies doubt that ``image-based representation learning provide limited value relative to a well-tuned baseline with image augmentations''~\shortcite{chen2022empirical}.
% DrQ~\shortcite{yarats2021image}, DrQ-v2~\shortcite{yarats2022mastering}.

% \subsection{Pretraining Dynamics}

% "Lastly, there exist model-based approaches that
% utilizes offline data to learn model dynamics which in tern accelerates imitation (Chang et al., 2021;
% Rafailov et al., 2021)"~\shortcite{yang2022trail}.

\section{Towards Generalist Agents with RL}\label{sec:generalist}

So far we have discussed online and offline scenarios that are generally restricted to a single modality and single environment.
Recently, there is a surge of interest in building a single \textit{generalist model}~\shortcite{reed2022generalist,lee2022multi,fan2022minedojo} to handle tasks in different environments across different modalities.
To enable the agent to learn from and adapt to various open-ended tasks, it is desired to leverage considerable prior knowledge in different forms such as visual perception and language understanding.
Intuitively, the aim is to bridge the worlds of RL and other fields of machine learning, combining previous success together to build a large decision-making model capable of a diverse set of tasks.
In this section, we look at various considerations for handling data and tasks from different modalities to acquire useful prior knowledge.

\subsection{Visual Pretraining}

Perception is an unavoidable prerequisite for real-world applications.
With an increased number of image-based decision-making tasks, pretrained visual encoders that were exposed to a wide distribution of images can provide RL agents with robust and resilient representations as a basis to learn optimal policies.

The field of computer vision has seen tremendous progress in pretraining visual encoders from large-scale image datasets~\shortcite{deng2009imagenet} and video corpora~\shortcite{abu2016youtube}.
Given that these data are cheap to access, several works have explored the use of pretrained visual encoders on large-scale image datasets as means to improve the generalization and sample efficiency of RL agents.
\shortciteA{shah2021rrl} equip standard deep RL algorithms with  ResNet encoders pretrained on ImageNet and observe that the pretrained representations lead to impressive performances in Adroit~\shortcite{rajeswaran2017learning} but struggle in the DeepMind control suite~\shortcite{tassa2018deepmind} due to large domain gap.
\shortciteA{parisi2022unsurprising} further investigate various design choices including datasets, augmentations, and layers, and report positive results on all four considered control tasks.
\shortciteA{trauble2022the} conduct a large-scale study on how different properties of pretrained VAE-based embeddings affect out-of-distribution generalization, concluding that some of them (e.g., the GS metric~\shortcite{dittadi2021on}) can be good proxy metrics to predict generalization performance.

Instead of extracting visual information from static image datasets, another intriguing direction is to capture temporal relations from unlabeled videos.
\shortciteA{sermanet2018time} design a self-supervised approach to learning temporal variance and multi-view invariance on multi-view video data.
\shortciteA{xiao2022masked} empirically find that, without exploiting temporal information, in-the-wild images collected from YouTube or Egocentric videos lead to better self-supervised representations for manipulation tasks that ImageNet images.
\shortciteA{seo2022reinforcement} introduce a two-phase learning framework, which first learns useful representations via generative pretraining on videos and then uses the pretrained model for learning action-conditional world models.
\shortciteA{baker2022video} successfully extract behavioral priors from internet-scale videos with an inverse dynamics model to uncover the underlying actions followed by behavior cloning, finding that the pretrained model exhibits impressive zero-shot capabilities and finetuning results for playing Minecraft.
\shortciteA{zhang2022learning} also leverage inverse dynamics models to predict action labels from action-free videos, upon which a new contrastive learning framework is proposed to pretrain action-conditioned policies.

% \zhihui{The following two are discussed in MineDojo.}
% R3M~\shortcite{nair2022r3m} shows visual representations pretrained on the Ego4D human videos~\shortcite{grauman2022ego4d} can boost data efficiency for downstream robotic manipulation tasks.
% CLIPort~\shortcite{shridhar2022cliport} combines the pretrained CLIP model~\shortcite{radford2021learning} with spatial reasoning frameworks to enable fine-grained manipulation based on language instructions.

\subsection{Natural Language Pretraining}

Human beings are not only able to perceive the visual world through their eyes, but understand high-level natural language instructions and ground the rich knowledge from texts to complete tasks.
In this vein, there has been a long history of how to connect language and actions~\shortcite{kollar2010toward,tellex2011understanding}.
Especially due to the rapid development of large language models (LLMs)~\shortcite{brown2020language,chowdhery2022palm} that exhibit great capability of encoding semantic knowledge, it appears to be a promising direction to leverage advanced LLMs as generic computation engines to facilitate decision making~\shortcite{lu2021pretrained}.

\subsubsection{Language-conditioned Policy Learning}

To extract and harness the knowledge of well-informed pretrained LLMs, a feasible solution is to condition the policies on text descriptions processed by LLMs.
% Pretrained LLMs can serve as a brain whereas the agent deployed in the environment plays the rule of hands and eyes.
% \shortcite{huang2022language} use prompt engineering to ground high-level tasks to a chosen set of admissible actions.
This kind of language-conditioned policy learning could be extremely useful for robotic tasks where high-level language instructions are available. 
For example, \shortciteA{ahn2022can} use pretrained LLMs to split high-level instructions into sub-tasks via prompt engineering for grounding value functions in real-world robotic tasks.
\shortciteA{huang2022inner} further enable grounded closed-loop feedback generated by additional perception models as the source of corrections for LLMs' predictions.
\shortciteA{tam2022semantic} instead consider effective exploration in 3D environments, showing that pretrained representations from vision-language models~\shortcite{radford2021learning} form a semantically meaningful state space for curiosity-driven intrinsic rewards.
\shortciteA{mahmoudieh2022zero} also connect reward specification to vision-language supervision, introducing a framework that leverages text descriptions and pixel observations to produce reward signals.

\subsubsection{Policy Initialization}

Recent advances bridge the gap between reinforcement learning and sequential modeling~\shortcite{chen2021decision,janner2021offline,zheng2022online,furuta2022generalized}, opening up opportunities to borrow sequential models to RL tasks.
Despite the clear distinction, pretrained LLMs could arguably provide reusable knowledge via weight initialization.
\shortciteA{reid2022can} investigate whether pretrained LLMs can provide good weight initialization for Transformer-based offline RL models, and conclude with very positive results.
\shortciteA{li2022pre} also demonstrate that pretrained LLMs can be used to initialize policies and facilitate behavior cloning as well as online reinforcement learning for embodied tasks.
They also suggest using sequential input representations and fintuning the pretrained weights for better generalization.

% \zhihui{The following are discussed in Pre-Trained Language Models for
% Interactive Decision-Making.}
% \shortcite{huang2022language}.
% \shortcite{lu2021pretrained}.
% \shortcite{sharma2021skill}.

\subsection{Multi-task and Multi-modal Pretraining}

With recent advances in building powerful sequence models to handle different modalities and tasks~\shortcite{lu2019vilbert,jaegle2022perceiver,wang2022image}, the wave of using large general-propose models~\shortcite{bommasani2021opportunities} has been sweeping through the field of supervised learning.
The key ingredient is Transformer~\shortcite{vaswani2017attention}, a highly capable neural architecture built on the self-attention mechanism~\shortcite{bahdanau2014neural} that excels at capturing long-range dependencies in sequential data.
Due to its strong generality where various tasks in different domains can be formulated as sequence modeling, Transformer is believed to be a unified architecture for developing foundation models~\shortcite{bommasani2021opportunities}.
% has been proven to be effective for text, on a wide range of tasks~\shortcite{devlin2018bert,dosovitskiy2021an,liu2020mockingjay,yin2020tabert}.

% Decision transformers~\cite{chen2021decision,janner2021offline,zheng2022online,furuta2022generalized} based on upside-down RL~\cite{srivastava2019training,kumar2019reward,schmidhuber2019reinforcement}.

Recently, Transformer-based architectures have also been extended to the field of offline RL~\shortcite{chen2021decision,janner2021offline} and then online RL~\shortcite{zheng2022online}, in which the agent is trained auto-regressively in a supervised manner via likelihood maximization.
This opens up the possibility of replicating previous success achieved with Transformer in the field of supervised learning.
Specifically, it is expected that by combining large-scale data, open-ended objectives, and Transformer-based architectures, we are ready to build general-purpose decision-making agents that are capable of various downstream tasks in different environments.

Pioneering work in this direction is Gato~\shortcite{reed2022generalist}, a generalist agent trained on various tasks from control environments, vision datasets, and language datasets in a supervised manner.
To handle multi-task and multi-modal data, Gato uses demonstrations as prompt sequences~\shortcite{wei2021finetuned} at inference time.
\shortciteA{lee2022multi} extend Decision Transformer~\shortcite{chen2021decision} to train a generalist agent called Multi-Game DT that can play 41 Atari games simultaneously.
Both Gato and Multi-Game DT show impressive scaling law properties.
\shortciteA{fan2022minedojo} make use of large-scale multi-modal data from YouTube videos, Wikipedia pages, and Reddit posts to train an agent able to solve various tasks in Minecraft.
To provide dense reward signals, a pretrained vision-language model based on CLIP~\shortcite{radford2021learning} is introduced as a proxy of human evaluation.
% "propose
% a novel agent learning algorithm that leverages large pre-trained video-language
% models as a learned reward function".

\subsection{Challenges \& Future Directions}

In spite of some promising results, how generalist models benefit from multi-modal and multi-task data remains unclear.
More specifically, these models might suffer from detrimental gradient interference~\shortcite{yu2020gradient} between modalities and tasks due to the incurred optimization challenges.
To mitigate this issue, it is desired to incorporate more analysis tools for optimization landscapes~\shortcite{goodfellow2014qualitatively} and gradients~\shortcite{yu2020gradient} to tease out the precise principles.

Another compelling direction is to compose separate pretrained models (e.g., GPT-3~\shortcite{brown2020language} and CLIP~\shortcite{radford2021learning}) together.
By leveraging expert knowledge from different models, this kind of framework can solve complex multi-modal tasks~\shortcite{li2022consensuscomposing}.
% \zhihui{Come back later.}

% Reward signals are used.

% \zhihui{
% "Research on what approaches to goal specification are both tractable for policy optimisation and useful for real-world scenarios
% would be beneficial, as there is likely a trade-off between these two desirable attributes. Hill et al. [142], Lynch and
% Sermanet [143] are good examples of investigating natural language as goal specification, utilising pretrained models to
% improve generalisation, and we look forward to seeing more work in this area." (generalisation survey)
% }

% of trajectories in the offline dataset.

% "Embodied agent learning? What do we want? Open-ended objectives, massively multitask, and world knowledge.
% How? open-ended environment, internet-scale knowledge, and foundation models"~minedojo talk

\section{Task Adaptation}\label{sec:downstream}

While pretraining on unsupervised experiences can result in rich transferable knowledge, it remains challenging to adapt the knowledge to downstream tasks in which reward signals are exposed.
In this section, we discuss briefly various considerations for downstream task adaptation.
We limit the scope to online adaptation, while adaptation with offline RL or imitation learning is also feasible~\shortcite{yang2021representation}.

In online task adaptation, a pretrained model is given, which can be composed of various components such as policies and representations, together with a target MDP that can interact with.
% Similar to the cases for supervised learning, adaptation in RL can be of .
Given that pretraining could result in different forms of knowledge, it brings difficulties to designing principled adaptation techniques.
Nevertheless, considerable efforts have been made to study this aspect.
% Task adaptation for RL is complicated by several factors.
% First of all, pretraining can result in different forms of prior knowledge.
% This brings difficulties to design principled adaptation techniques.
% Next, "the transfer of prior knowledge in the context of an MDP, instead of a stationary data domain as in supervised learning, means that observations from different RL domains are not identically and independently distributed (i.i.d); as such, it is even less likely that observations in different reinforcement learning tasks will be similarly distributed".
% Finally, "because the transition dynamics and reward functions of RL tasks are usually not known in advance and
% must be sampled as a through a process, it is difficult
% if not impossible to determine the similarity of two
% different tasks".

% "The benefits of unsupervised pre-
% training are typically evaluated by their ability to enable efficient transfer to previously unseen reward
% functions [28]"~\shortcite{campos2021beyond}.
% After pretraining, one can adapt to a new task with a short phase of online interaction where rewards are provided.
% The scenario is called \textit{few-shot adaptation}\zhihui{~\shortcite{mitchell2021offline,xu2022prompt}}.
% One can also consider \textit{zero-shot adaptation} (\zhihui{with Model-based RL~\shortcite{Sharma2020Dynamics-Aware,sekar2020planning}} where no task-specific experience is given and the pretrained agent is directly evaluated in the downstream task.

% \subsection{Few-shot Adaptation}

\subsection{Representation Transfer}

In the field of supervised learning, recent advances~\shortcite{devlin2018bert,he2020momentum,chen2020simple} have demonstrated that good representations can be pretrained on large-scale unlabeled dataset, as evidenced by their impressive downstream performances.
The most common practice is to freeze the weights of the pretrained feature encoder and train a randomly initialized task-specific network on top of that during adaptation.
The success of this paradigm is essentially based on the promise that related tasks can usually be solved using similar representations.

For RL, it has been shown that directly reusing pretrained task-agnostic representations can significantly improve sample efficiency on downstream tasks.
For instance, \shortciteA{schwarzer2021pretraining} conduct experiments on the Atari 100K benchmark and find that frozen representations pretrained on exploratory offline data already form a basis of data-efficient RL.
This success also extends to the cases where domain discrepancy exists between upstream and downstream tasks~\shortcite{shah2021rrl,parisi2022unsurprising}.
However, the issue of \textit{negative transfer} in the face of domain discrepancy might be exacerbated for RL due to its complexity~\shortcite{shah2021rrl}.

% Recently, it has been shown that representation transfer provably accelerate RL~\shortcite{agarwal2022provable}.
% However, the study only consider supervise multi-task pretraining, and unsupervised pretraining remains unclear.

% Main advantage: decoupled structure help scaling up; directly using pretrained visual encoders.

When adapting to tasks that have the same environment dynamics as that of the upstream task(s), successor features~\shortcite{barreto2017successor} can be a powerful tool to aid task adaptation.
The framework of successor features is based on the following decomposition of reward functions:

\begin{equation}\label{eq:successor_feature}
    r\left(s, a, s^{\prime}\right)=\phi\left(s, a, s^{\prime}\right)^{\top} w,
\end{equation}
where $\phi\left(s, a, s^{\prime}\right) \in \mathbb{R}^d$ represents features of transition $\left(s, a, s^{\prime}\right)$ and $w \in \mathbb{R}^d$ encodes reward-specifying weights.
This leads to a representation of the value function that decouples the dynamics of the environment from the rewards:

\begin{equation*}
    Q^\pi(s, a) =\mathbb{E}_{s_t=s,
    a_t=a}\left[\sum_{i=t}^{\infty} \gamma^{i-t} \phi\left(s_{i+1}, a_{i+1}, s_{i+1}^{\prime}\right)\right]^\top w
    = \psi^\pi(s, a)^\top w,
\end{equation*}
where we call $\psi^\pi(s, a)$ the \textit{successor features} of $(s, a)$ under $\pi$.
Intuitively, $\psi^\pi$ summarizes the dynamics induced by $\pi$ and has been studied within the framework of online pretraining~\shortcite{Hansen2020Fast,liu2021aps} by combining with skill discovery approaches to implicitly learn controllable successor features $\psi^\pi(s, a)$.
Given a learned $\psi^\pi(s, a)$, the problem of task adaptation reduces to a linear regression derived from Equation~\ref{eq:successor_feature}.

% \zhihui{Similar to SFs~\shortcite{touati2021learning}.}

\subsection{Policy Transfer}\label{sec:policy_transfer}

A compelling alternative for task adaptation is to transfer learned behaviors.
As discussed in previous sections, existing work has explored how to pretrain primitive skills that can be reused to face new tasks or a single exploratory policy that facilitates exploration at the beginning of task adaptation.
The differences in pretrained behaviors result in different adaptation strategies.

To achieve high rewards on the downstream task with skill-conditioned policy $\cpolicy$, a straightforward strategy is to simply choose the skill $z$ with the best outcome and further enhance it with finetuning.
However, a single best-performing skill can not fulfill its potential.
To better combine diverse skills for task solving, one can view them from the perspective of \textit{hierarchical RL}~\shortcite{barto2003recent,kulkarni2016hierarchical}.
In hierarchical RL, the decision-making task is typically decomposed into a two-level hierarchy, where a meta-controller $\controller$ decides which low-level policy to use for task solving, depending on the current state.
This hierarchical scheme is agnostic to how the low-level policies are learned.
Therefore, it is sufficient to train a meta-controller on top of the discovered skills, which has been proven effective for few-shot adaptation~\shortcite{hakhamaneshi2022hierarchical} and zero-shot adaptation~\shortcite{Sharma2020Dynamics-Aware}.

Exploratory policies, as another form of prior knowledge, benefit downstream tasks in a different way.
Due to the importance of exploration, exploratory policies can provide good initialization for the agent to gather diverse experiences and reach high-rewarding states.
For example, \shortciteA{campos2021beyond} validate the
effectiveness of transferring exploratory policies trained by curiosity-driven approaches, in particular for domains that require structured exploration.

While it is always feasible to finetune pretrained policies, considerations should be taken in order to prevent \textit{catastrophic forgetting} when learning in the downstream task.
Catastrophic forgetting refers to the tendency
of neural networks to disregard their previously obtained knowledge when new information is acquired.
To mitigate this issue, one might apply knowledge distillation-like regularization together with RL objectives~\shortcite{schmitt2018kickstarting}:

\begin{equation*}
    \mathcal{L}_{\text{KD}}=H\left(\hat{\pi}(a \mid s) \| \pi_\theta(a \mid s)\right),
\end{equation*}
where $H$ is cross entropy and $\hat{\pi}$ is the teacher policy.
We refer the reader to \shortciteA{khetarpal2020towards} for more discussions on catastrophic forgetting in reinforcement learning.
% the pre-trained
% policy, thus prematurely disregarding its exploratory behavior.

% Fine-tuning a pre-trained exploratory policy can be very beneficial, as the agent will gather diverse experiences more quickly than randomly initialized agents or .
% \zhihui{Beyond transfer}.

% \input{tab/downstream_setup}

% \subsection{Speed up in the Same Tasks}

% How about comparing with data augmentation approaches?
% If we only care about transfer within the same environment, differences between two framework: pretraining and joint learning~\shortcite{chen2022empirical,li2022does}.

% Once the reward function is specified.
% SGI~\cite{schwarzer2021pretraining}.

% Imitation learning, online RL, and offline RL can all take advantages~\cite{ajay2021opal,yang2021representation}.

\subsection{Challenges \& Future Directions}

\paragraph{Parameter Efficiency.}
Despite that existing pretrained models for RL have much fewer parameters as compared with those in the field of supervised learning, the issue of parameter efficiency is still important with the ever-increasing number of model parameters.
More concretely, it is desired to design \textit{parameter-efficient transfer learning} that updates only a small fraction of parameters while keeping most of the pretrained parameters intact.
It has been actively studied in natural language processing~\shortcite{he2021towards} with solutions like adding small neural modules as adapters~\shortcite{houlsby2019parameter} and prepending learnable prefix tokens as soft prompts~\shortcite{lester2021power}.
Built on these techniques, several efforts have been made to enable parameter-efficient transfer with prompting~\shortcite{xu2022prompt,reed2022generalist}, which we believe has a large room to improve with tailored methods.

% \paragraph{Safety.}
% "For
% instance, the environment may have dangerous states or harmful objects that the agent should avoid, even though they would make it curious during pre-training"~\shortcite{parisi2021interesting}.

\paragraph{Domain adaptation.}
In this section, we mainly consider task adaptation where unseen tasks are given in the same environment.
A more challenging but practical scenario is domain adaptation.
In domain adaptation, there exist environmental shifts between the upstream and downstream tasks.
Despite that these environmental shifts are commonly seen in real-world applications, it remains a challenging problem to transfer across different domains~\shortcite{eysenbach2020off,huang2021adarl}.
However, we believe that this direction will rapidly evolve by bringing related techniques from supervised learning to reinforcement learning.
% ``Pretraining for rapid adaptation to new games has not been explored widely on Atari games despite being a natural and well-motivated task due to its relevance to how humans transfer knowledge to new games.'' (multi-game)

% \paragraph{Measuring Task Similarity.}

% Measuring task similarity to avoid negative transfer.

\paragraph{Continually-developed models.}
For practical applications, we can take a step forward and consider building large pretrained models continually to support added features (e.g., a modified action space, more powerful architectures, etc).
While such consideration was already underway during the development of large-scale RL models~\shortcite{berner2019dota}, it requires a more principled way of combining updates into RL models.
We refer the reader to recent work in this direction in the field of supervised learning~\shortcite{raffel2021call} and reinforcement learning~\shortcite{campos2021beyond,agarwal2022beyond}.
% \zhihui{"the same neural network
% architecture needs to be used for both the pre-trained and the downstream policies (for
% instance, if the pre-trained policy was trained using a policy-based method, it might not be possible
% to fine-tune it using a value-based approach)"~\shortcite{campos2021beyond}.}
% "Moreover, the same neural network
% architecture needs to be used for both the pre-trained and the downstream policies, which in practice
% also imposes a limitation on the type of RL methods that can be employed in the adaptation stage (for
% instance, if the pre-trained policy was trained using a policy-based method, it might not be possible
% to fine-tune it using a value-based approach)."~\shortcite{campos2021beyond}
% This is partially addressed with \shortcite{agarwal2022beyond}.
% \input{sec/6-theoretical}
\section{Conclusions and Future Perspectives}\label{sec:conclude}

In this section, we conclude this survey and highlight several future directions which we believe will be important topics for future work.

This paper introduces pretraining in deep RL by discussing recent trends to obtain general prior knowledge for decision-making.
In contrast to its supervised learning counterpart, pretraining faces a variety of challenges unique to RL.
In this survey, we present several promising research directions to tackle these challenges and we believe this field will evolve rapidly in the coming years.

There are still several open questions that are important and remain to be addressed.
% Advocate for open access of checkpoints~\shortcite{agarwal2022beyond,gogianu2022agents}.

% \paragraph{Applications.}
% Currently we have seen applications mainly in simulated environments.

% Robotics.
% Games.
% Beyond that.

\paragraph{Benchmarks and evaluation metrics.}
Evaluation serves as a means for comparing various methods and driving further improvements.
In the field of natural language processing, GLUE~\shortcite{wang2018glue} is a widely used benchmark to evaluate the performance of models across various natural language understanding tasks.
% \zhihui{Here we list several famous benchmarks in NLP/CV.}
Recently, there has been a surge of research on improving evaluation for RL in terms of evaluation metrics~\shortcite{agarwal2021deep} and benchmark datasets~\shortcite{cobbe2020leveraging}.
To the best of our knowledge, URLB~\shortcite{laskin2020curl} is the only benchmark for pretraining in deep RL.
It presents a unified evaluation protocol for online pretraining based on the DeepMind control suite~\shortcite{tassa2018deepmind}.
However, a principled evaluation framework for offline pretraining and generalist pretraining is still missing.
We expect existing offline RL benchmarks like D4RL~\shortcite{fu2020d4rl} and RL Unplugged~\shortcite{gulcehre2020rl} can serve as the basis for developing pretraining benchmarks, but more challenging tasks should better illustrate the value of pretraining.
% Until \zhihui{XXX}, there is little effort made in this vein (URLB~\cite{laskin2021urlb}).
    
\paragraph{Architecture.}
As discussed in previous sections, there has been a surge of leveraging large transformers for RL tasks.
We expect other recent advances in model architecture can bring more improvements.
For example, \shortciteA{mustafa2022multimodal} learn large sparse models with a mixture of experts that simultaneously handle images and text with modality-agnostic routing.
This has the promise to solve complex tasks at scale.
Besides, one can also rethink existing architectures that have the potential to support large-scale pretraining (e.g., Progress Neural Networks~\shortcite{rusu2016progressive}).
% A unified picture (across modalities/tasks).

% MetaMorph~\cite{gupta2022metamorph}.
    
\paragraph{Multi-agent RL.}
Multi-agent RL~\shortcite{zhang2021multi} is an important sub-field of RL.
Extending existing pretraining techniques to the multi-agent scenario is non-trivial.
Multi-agent RL typically requires socially desirable behaviors~\shortcite{ndousse2021emergent} and representations~\shortcite{li2021celebrating}.
To the best of our knowledge, \shortciteA{meng2021offline} present the only effort in pretraining multi-agent RL with supervision.
How to enable unsupervised pretraining for multi-agent RL remains unclear, which we believe is a promising research direction.

\paragraph{Theoretical results.}
For RL, the significant gap between theory and practice has been a long-standing problem, and bringing large-scale pretraining to RL may exacerbate this even further.
Fortunately, recent theoretical studies have made efforts in terms of representation transfer~\shortcite{agarwal2022provable} and skill-conditioned policy transfer~\shortcite{eysenbach2021information}.
Increasing
focus on theoretical results is likely to have profound effects on the development of more advanced pretraining methods.`
% \paragraph{More.}
% A more general paradigm for learning primitives (like prompt-based learning), which is non-trivial as decision-making tasks are extremely unstructured.

\newpage

\vskip 0.2in
\bibliography{sample}
\bibliographystyle{newtheapa}

\end{document}